\documentclass{article}

\usepackage{arxiv}
\usepackage[utf8]{inputenc} 
\usepackage[T1]{fontenc}    
\usepackage{hyperref}       
\usepackage{url}            
\usepackage{booktabs}       
\usepackage{amsfonts}       
\usepackage{nicefrac}       
\usepackage{microtype}      
\usepackage{graphicx}
\usepackage{amsmath}
\usepackage{subcaption}
\def\style{naturemag}
\usepackage{comment}
\usepackage{xcolor}
\usepackage{placeins}

\usepackage[ruled,vlined]{algorithm2e}

\title{Automated Behavioral Analysis Using Instance Segmentation}

\author{
  Chen Yang \\
  Department of Psychology\\
  Cornell University\\
  Ithaca, NY 14853 \\
  \texttt{cy384@cornell.edu} \\
\And
  Jeremy Forest \\
  Department of Psychology\\
  Cornell University\\
  Ithaca, NY 14853 \\
  \texttt{jerem.forest@gmail.com} \\
\And
  Matthew Einhorn\\
  Department of Psychology\\
  Cornell University\\
  Ithaca, NY 14853 \\
  \texttt{me263@cornell.edu} \\
\And
  Thomas A. Cleland \\
  Department of Psychology\\
  Cornell University\\
  Ithaca, NY 14853 \\
  \texttt{tac29@cornell.edu} \\
}

\begin{document}
\maketitle
\begin{abstract}
Animal behavior analysis plays a crucial role in various fields, such as life science and biomedical research. However, the scarcity of available data and the high cost associated with obtaining a large number of labeled datasets pose significant challenges. In this research, we propose a novel approach that leverages instance segmentation-based transfer learning to address these issues. By capitalizing on fine-tuning the classification head of the instance segmentation network, we enable the tracking of multiple animals and facilitate behavior analysis in laboratory-recorded videos. To demonstrate the effectiveness of our method, we conducted a series of experiments, revealing that our approach achieves exceptional performance levels, comparable to human capabilities, across a diverse range of animal behavior analysis tasks. Moreover, we emphasize the practicality of our solution, as it requires only a small number of labeled images for training. To facilitate the adoption and further development of our method, we have developed an open-source implementation named Annolid (An annotation and instance segmentation-based multiple animal tracking and behavior analysis package). The codebase is publicly available on GitHub at https://github.com/cplab/annolid. This resource serves as a valuable asset for researchers and practitioners interested in advancing animal behavior analysis through state-of-the-art techniques.

\end{abstract}


\section{Introduction}
The primary objective of animal behavior analysis is to identify every instance of a specific behavior within a video and determine its precise spatial and temporal localization. This analytical process finds practical applications in various fields of biomedical research and notably contributes to a better understanding of the neural mechanisms involved in both healthy and diseased behaviors. Due to the diverse array of laboratory environments and behavioral apparatuses available (e.g., open field arenas, Morris water mazes, three-chamber assays), accurately quantifying subtle variations in behavior, such as grooming, can pose significant challenges. Moreover, manually observing videos to extract metrics related to the behavior of interest not only demands extensive amounts of researchers' time but also fails to accommodate the scale of recorded behavioral studies. This manual approach also introduces a high degree of subjectivity to the obtained results. Automated animal pose estimation and tracking have witnessed remarkable advancements in recent years, thanks to the rapid progress in deep learning approaches \cite{mathis2020primer,Pereira331181,Pereira2020.08.31.276246,lauer2021multi}. These developments have been instrumental in revolutionizing the field. The fundamental idea revolves around training deep learning networks using annotated keypoint data that effectively represent animals. Armed with this training, the network exhibits exceptional proficiency in accurately estimating animal poses when presented with new video footage. The field of animal behavior analysis encompasses a wide range of applications \cite{Arac2019DeepBehavior,mathis2020primer,segalin2020mouse,Pereira331181,Pereira2020.08.31.276246,sun2021multi,lauer2021multi}. For instance, \cite{Bohnslav2020.09.24.312504} developed DeepEthogram, an approach that utilizes convolutional neural network models to process videos and extract motion features from individual frames. The resulting models demonstrate remarkable accuracy of over 90\% in classifying user-defined behaviors within fly and mouse videos, even on a single-frame basis. The existing pose estimation techniques have a noteworthy drawback: they primarily rely on key point representation, which lacks the flexibility to directly model animal behaviors. Additionally, the classification at the frame level often lacks the necessary precision to accurately locate animals and their behaviors in space. The demand for experiments that can measure multiple animals and their interactions with the environment, including specific objects within it, is increasing. Current pose estimation-based methods do not address certain tasks such as tracking and scoring urine deposition based on thermal camera images. Existing behavioral tracking software often lacks or provides inaccurate tracking of multiple objects or animals. Over the past few years, significant strides have been made in enhancing the accuracy and effectiveness of object detection and instance segmentation. These notable advancements can be largely attributed to the remarkable progress in deep learning techniques \cite{he2017mask,kirillov2019pointrend,yolact-plus-arxiv2019,yolact-iccv2019}. When it comes to tasks involving the assessment of an animal's interaction with objects or the measurement of its investigation area or perimeter, instance segmentation methods have emerged as a more appropriate and well-suited approach.

Multi-Object Tracking (MOT) is a computer vision task that estimates trajectories for objects of interest in videos \cite{zhou2020tracking,sun2018deep}. Many Multi-Object Tracking methods have two stages: 1) object detection and 2) data association \cite{ciaparrone2020deep}. The first stage detects objects in each frame while the second stage associates the corresponding detections across frames to obtain tracks. Nowadays, Tracking-by-detection is dominating the multiple object tracking pipelines by object detection followed by temporal association or re-identification (re-ID) \cite{zhang2020fairmot}. Zhang et al. (2020) shows that the re-ID accuracy is mainly depending on the detection task and is not fairly learned which causes many identity switches for multiple object tracking. Most of the appearance features are not robust to object deformations, occlusions, and illumination variations and they also failed to discriminate objects with high similarity. The motion models use linear motion models with constant velocity assumptions or non-linearity motion models to predict the object locations in future frames. However, these motion models failed to deal with occlusions and interactions well and the models that tried to combine motion and appearance modeling are having a hard time to archive a balance in the real world. In the behavior analysis domain, the body part's motion is the output of the neural electrical activity. Annotating and analyzing animal behaviors from lengthy recorded videos are time-consuming tasks for scientists. With the advance of recent machine learning methods, many toolkits have been developed to automate behavior analysis tasks to track animal centroids and recognize user-defined behaviors, such as rearing, grooming, or fighting \cite{Bohnslav2020.09.24.312504}. Bohnslav et al. (2020) model takes each frame to predict whether each behavior is present or absent without the instance-level information. This approach does not know where the behaviors occurred and will not work if there are multiple animals performing different behaviors. Animal behavior analysis has benefited from the application of Multiple Object Tracking (MOT) techniques, which enable the tracking and analysis of individual animals within a group. However, there are several challenges and issues that must be considered when using MOT for animal behavior analysis. One challenge is ambiguity. Animals within a group often have similar appearances, making it difficult for trackers to distinguish between them accurately. This ambiguity can result in incorrect tracking and identity switches, compromising the reliability of the data. Occlusions pose another challenge. Animals may obstruct each other's movements, leading to incomplete or inaccurate tracking. This becomes particularly problematic when animals move unpredictably or when they are in close proximity to one another. Scale and perspective present additional hurdles. As animals move farther away from the camera, their size and shape may change, making it challenging for trackers to maintain accurate identification. Lighting conditions can also impact tracking accuracy. Changes in lighting can affect the visibility of animals and make it difficult for trackers to distinguish between them accurately. Interactions and group dynamics further complicate the tracking process. Animals' behaviors and movements within a group are often complex and dynamic, making it challenging to identify and track individual animals. This complexity can result in incomplete or inaccurate tracking data. Additionally, the manual annotation of animal behavior data is time-consuming and requires significant expertise. This limitation hampers the scalability and reproducibility of animal behavior studies. In summary, while MOT has the potential to offer valuable insights into animal behavior, the challenges mentioned above need to be carefully considered and addressed to ensure accurate tracking data. Such tasks are not being addressed by current pose estimation software packages like DeepLabCut \cite{Mathisetal2018,Mathis_2020}. Occlusions and crowded scenes also remain highly challenging and additional steps are needed to use the body positions to classify behaviors. 

To address these challenges, a new approach called annolid has been introduced. Annolid is an annotation and segmentation-based multiple animal tracking and behavior analysis system. It aims to overcome the difficulties associated with tracking multiple animals by providing improved annotation and segmentation tools for accurate tracking and behavior analysis. Annolid can help mitigate these challenges and enhance the accuracy and efficiency of animal behavior analysis.The following are the major contributions of this study:
\begin{itemize}
\item Demonstrating the efficacy of utilizing detailed object representations, such as segmentation, to model behavioral patterns.
\item Introducing a unique identification system where each animal is labeled with an individual ID, which is used as the classification label for the instance segmentation network. This eliminates the need for an association step in multiple animal tracking.
\item Constructing a novel dataset specifically designed for multi-animal tracking.
\item Creating and making available the annolid package, an open-source toolset for annotating, training, and inferring segmentation networks.
\end{itemize}

\section{Methods}

We addressed the challenges by employing deep learning-based methods of instance segmentation, as demonstrated to be successful in computer vision tasks \cite{He_2017_ICCV}. Behavioral laboratory experiments, being less complex than real-world environments, allowed us to utilize a small number of labeled images per animal type and experimental settings, achieving state-of-the-art behavioral analysis. By leveraging pre-trained models from large datasets like ImageNet and COCO \cite{lin2014microsoft}, we employed transfer learning for animal behavior analysis with a limited number of labeled image examples.Our method utilizes the pre-trained Mask R-CNN model from Detectron2, which can be effectively trained on a custom dataset of approximately 200 labeled frames in just 3 hours using Colab or within 30 minutes on an Nvidia 1080Ti GPU. Additionally, we utilized GPU-based YOLACT models, which offer real-time prediction speeds of around 30 frames per second. These models enable us to track animals by segmenting the desired region or defined behavior and take advantage of over-fitting to achieve human-level performance with around 100 labeled frames. We evaluated the performance of our approach on various datasets containing freely moving animals, including mice and voles. In our study, we employed two strategies. Firstly, we fine-tuned a pre-trained instance segmentation network, Mask-RCNN \cite{he2017mask}, to achieve highly accurate predictions. Alternatively, we utilized faster instance segmentation networks called YOLACT \cite{yolact-iccv2019} and YOLACT++ \cite{yolact-plus-arxiv2019}, which provide real-time performance but sacrifice a bit of accuracy. These models simultaneously segmented and tracked animals in each video frame. To ensure accurate instance separation and prevent confusion between animals, we assigned a unique class label to each animal (e.g., animal\_1, animal\_2, ..., animal\_X). This labeling scheme enabled the network's classification head to learn distinct appearance features for different instances, minimizing ID switches or confusion among animals. Consequently, re-identifying animals across frames was avoided by associating the Regions of Interest (ROIs) throughout the video sequence. 

Introducing Annolid (Annotation and Annelid with segmentation), we presented a novel behavioral software that utilizes instance segmentation for multiple object tracking and behavior analysis. Leveraging transfer learning from pre-trained instance segmentation networks based on the COCO dataset \cite{lin2014microsoft}, Annolid automatically segments and tracks animals and objects while classifying user-defined behaviors. This approach significantly reduces the need for extensive labeling, requiring only a small number of labeled frames (approximately 100). Annolid offers a unified graphical user interface that simplifies dataset creation and visualization. It supports training instance segmentation models using popular methods such as Mask R-CNN \cite{He_2017_ICCV}, PointRend \cite{kirillov2019pointrend} implemented in Detectron2 \cite{wu2019detectron2}, YOLACT \cite{yolact-iccv2019}, and YOLACT++ \cite{yolact-plus-arxiv2019}. These models can also be used for real-time prediction on new data. However, it is important to note that annotating and training the model for each new video featuring an unseen animal is necessary. Our experimental results demonstrate that our approach achieves performance levels comparable to humans, even without a large set of labeled images or the need for separate recognition networks such as Siamese networks \cite{koch2015siamese} for face recognition. This is particularly valuable because data acquisition for specific animals is often limited in laboratories, and the cost associated with labeling a large number of training examples is high. Using Annolid is straightforward and follows a recommended sequence of steps:
\begin{table}[h]
\centering
\begin{tabular}{|c|p{0.7\linewidth}|}
\hline
\textbf{Step} & \textbf{Description} \\
\hline
1 & Selection of desired frames for labeling. \\
\hline
2 & Labeling the selected frames. \\
\hline
3 & Conversion and creation of the dataset in COCO format. \\
\hline
4 & Training the selected network. \\
\hline
5 & Scoring the video using the trained model. \\
\hline
6 & Performing statistical analysis on the tracking results. \\
\hline
\end{tabular}
\caption{Recommended sequence of steps using Annolid.}
\end{table}

\begin{figure}[h]
    \centering
    \includegraphics[width=\linewidth]{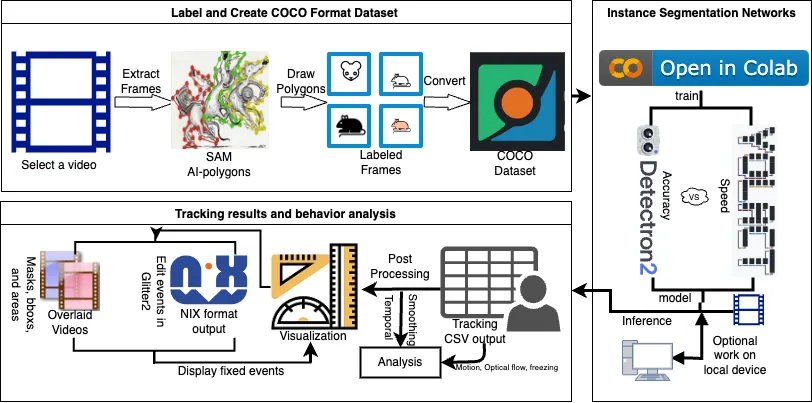}
    \caption{Overview of Annolid workflow}
    \label{fig:annolid_workflow}
\end{figure}

We have developed a user-friendly Graphical User Interface (GUI) based on the labelme \cite{labelme2016} tool to assist users throughout their workflow. In addition, Annolid offers users the flexibility to choose the instance segmentation method that best suits their datasets. They can opt for YOLACT, YOLACT++, Mask-RCNN, or PointRend. Our main contribution lies in adopting instance segmentation networks for tracking. We treat each instance as its own class frame by frame, eliminating the need for the association step required in traditional Multiple Object Tracking (MOT) methods. By doing so, Annolid achieves both instance segmentation and data association without the need for additional track matching. Furthermore, we have eliminated the requirement to maintain tracklets since we have precise information about the location and identities of objects in each frame. Figure \ref{fig:annolid_workflow} shows an overview of Annolid's workflow. 

\begin{figure}
    \centering
    \includegraphics[scale=0.45]{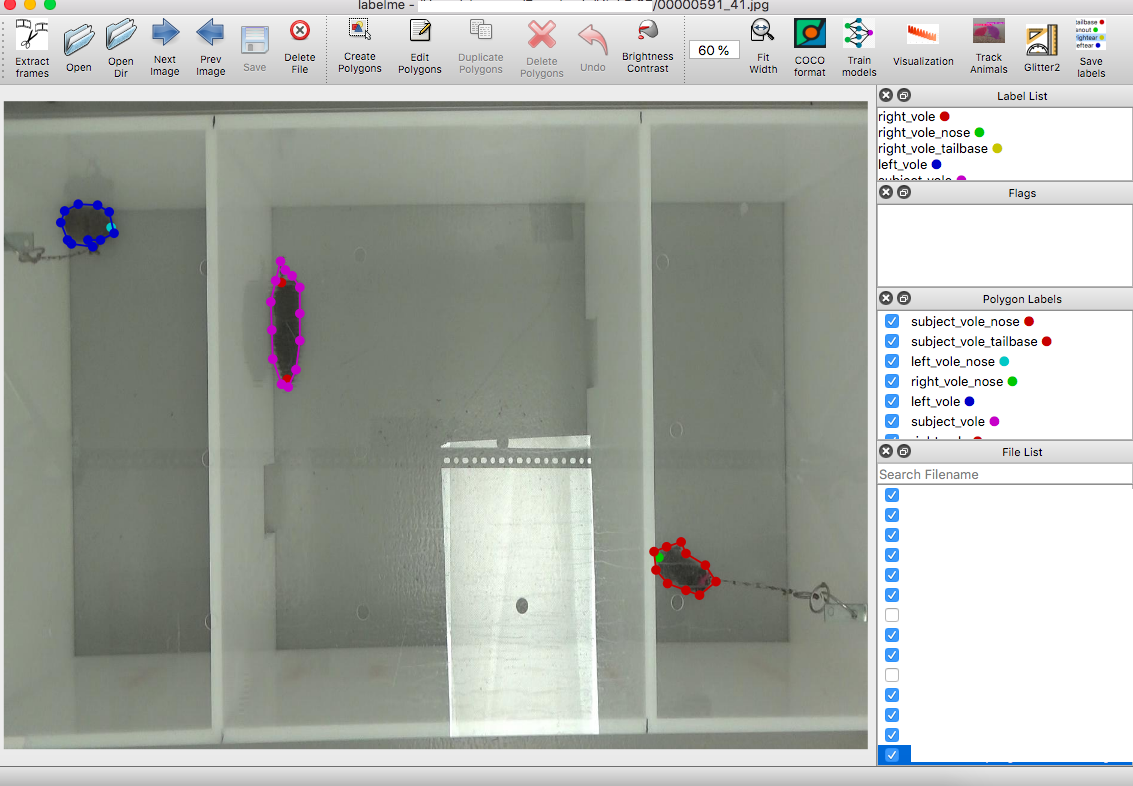}
    \caption{Overview of the Annolid GUI based on labelme\cite{labelme2016}.}
    \label{fig:annolid_gui}
\end{figure}

\begin{figure}[!htb]
\minipage{0.32\textwidth}
  \includegraphics[width=\linewidth]{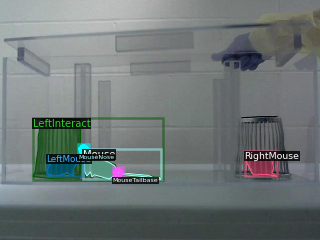}
  \label{fig:seg_anno1}
\endminipage\hfill
\minipage{0.32\textwidth}
  \includegraphics[width=\linewidth]{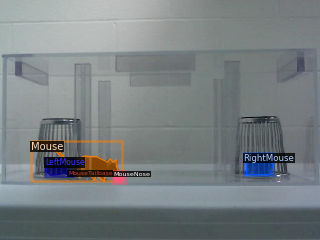}
  \label{fig:seg_anno2}
\endminipage\hfill
\minipage{0.32\textwidth}%
  \includegraphics[width=\linewidth]{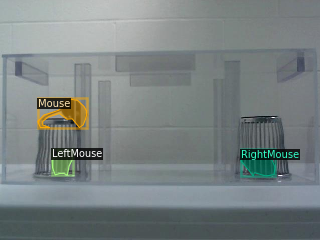}
  \label{fig:seg_anno3}
\endminipage
\caption{Example frames of mice with manually annotated masks.}
\label{fig:three_mice}
\end{figure}

\begin{figure}
    \centering
    \includegraphics[scale=0.45]{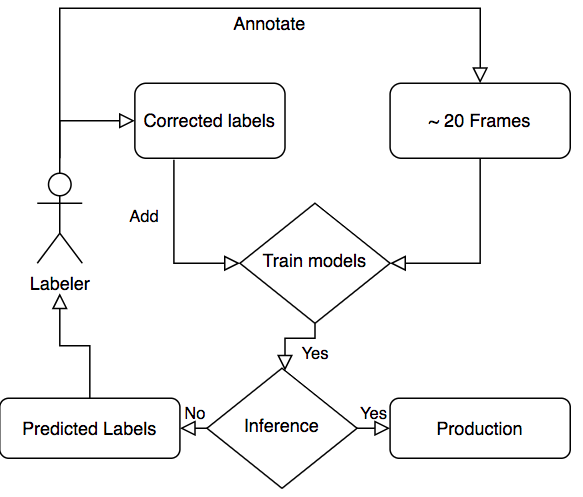}
    \caption{Human in the loop with model auto-labeling.}
    \label{fig:annolid_human_in_the_loop}
\end{figure}

\begin{figure}[!htb]
\minipage{0.32\textwidth}
  \includegraphics[width=\linewidth]{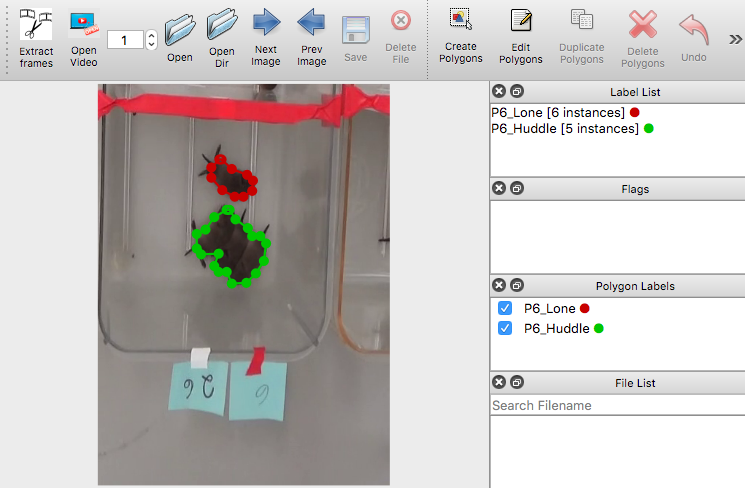}
  \label{fig:auto_polygons_1}
\endminipage\hfill
\minipage{0.32\textwidth}
  \includegraphics[width=\linewidth]{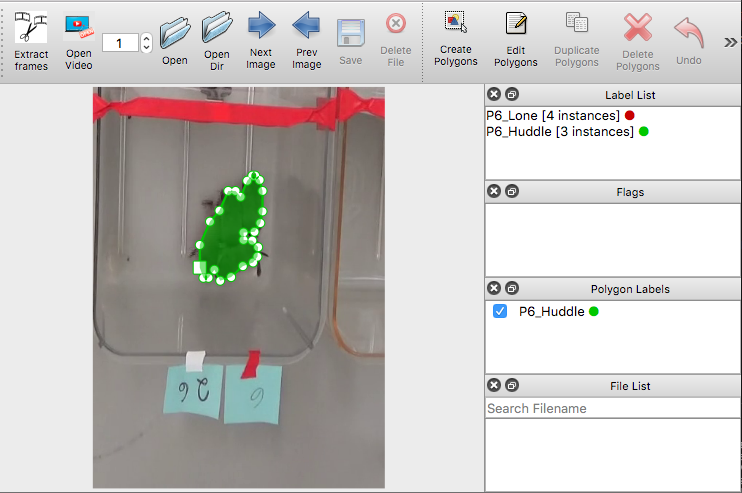}
  \label{fig:auto_polygons_2}
\endminipage\hfill
\minipage{0.32\textwidth}%
  \includegraphics[width=\linewidth]{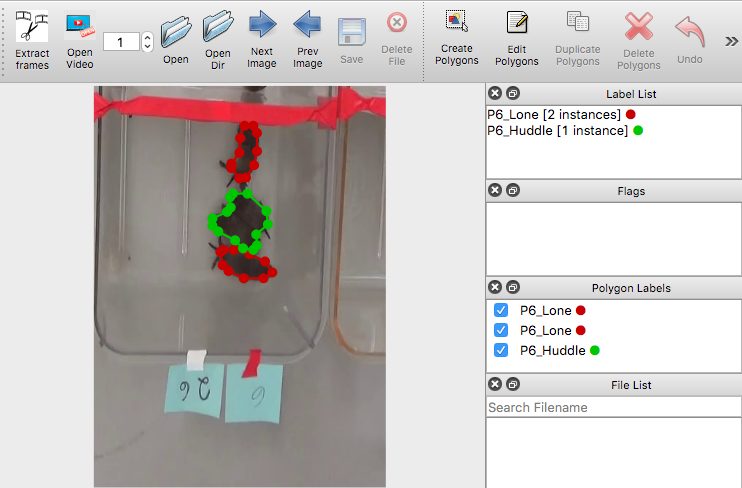}
  \label{fig:auto_polygons_3}
\endminipage

\caption{Example: Predicted polygons by a trained instance segmentation model. In this use case, each polygon represents either huddling or lone behavior. }
\end{figure}

\begin{figure}[!htb]
\minipage{0.32\textwidth}
  \includegraphics[width=\linewidth]{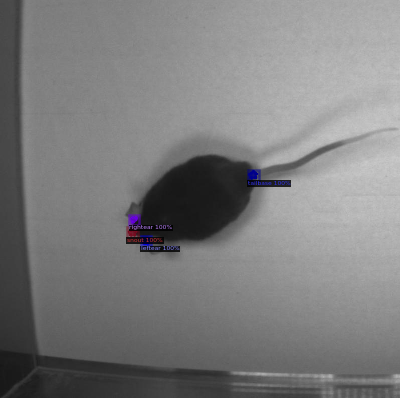}
  \label{fig:m8s1_test_1}
\endminipage\hfill
\minipage{0.32\textwidth}
  \includegraphics[width=\linewidth]{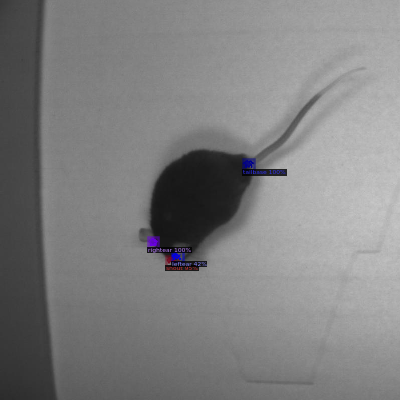}
  \label{fig:m8s1_test_3}
\endminipage\hfill
\minipage{0.32\textwidth}%
  \includegraphics[width=\linewidth]{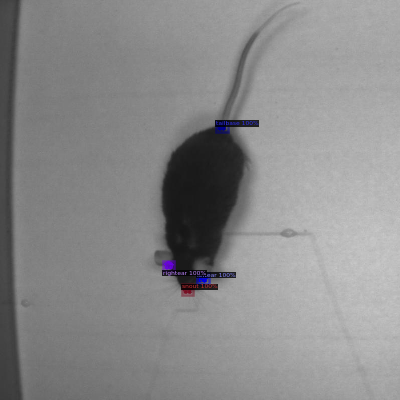}
  \label{fig:m8s1_test_4}
\endminipage
\caption{An example of predicted body parts on the dataset\cite{nath2019using}, treating pose estimation as a special instance segmentation case.}
\label{fig:pose_estimation}
\end{figure}

\subsection{Integration of the Segment Anything Model(SAM) into Annolid}
In this section, we discuss the seamless integration of the Segment Anything Model, introduced by Kirillov et al. in their 2023 paper \cite{kirillov2023segany}, into the Annolid platform. The Segment Anything model, developed and released by Meta research, has undergone extensive training with a vast dataset comprising more than one billion masks. This training has endowed the model with a high level of sophistication and the ability to accurately identify and delineate instances within images.One of the key objectives of integrating the Segment Anything model into Annolid is to simplify the instance labeling process for users. Traditionally, instance segmentation tasks can be time-consuming and labor-intensive, requiring meticulous manual annotation. Within the Annolid platform, users are presented with an intuitive interface that allows them to label instances with remarkable ease. Users need only designate a few positive and negative points within the image. These points serve as guidance for the Segment Anything model. It automatically generates precise masks, which are then processed into polygons outlining the identified instances within the image. These polygons can be further simplified using map generalization techniques commonly employed in geographic information systems (GIS), as implemented in \cite{Hugel_Simplification_2021}. This automation significantly reduces the time and effort typically associated with manual polygon drawing and instance delineation. While the automatic polygon generation is highly accurate, Annolid provides users with the flexibility to fine-tune the generated polygons. This feature ensures that users can achieve the level of precision required for their specific labeling tasks. Beyond instance delineation, users have the capability to assign class labels to the generated polygons based on their research requirements, even assigning behaviors directly as the polygon name. The integration of the SAM into Annolid promises a significant improvement in efficiency and accuracy in the labeling process.

\subsection{Classification head}
To distinguish between instances in a given video, we utilized the classification head of the instance segmentation model. Specifically, we labeled each instance as a distinct class (e.g., "mouse1" and "mouse2") across frames, instead of assigning a single class label such as "mouse". The unique identity assigned to each instance was then used as the training class label in the instance segmentation's classification branch. This approach ensures that the network is trained to classify instances correctly, even in frames that were not labeled. However, due to motion and occlusion, the same instance can disappear in some frames either fully or partially (e.g., due to body part occlusion). To account for this, the model should be given a set of frames representing different appearances of the same object throughout the video. Various methods are available for extracting frames for labeling, such as random sampling, optical flow, and keyframes. In our experiments, we used a random sampling method for selecting training frames to label, as it was faster.

\subsection{Datasets and Metrics}
In order to label and prepare datasets, we utilize our Annolid user interface, which is based on Labelme. This interface allows us to create polygons for instances and regions of interest for segmentation masks. These masks can be used for various purposes, such as identifying behavioral interactions or body parts. The interface generates Labelme ".json" files, which can be converted to COCO format datasets. These datasets are suitable for training instance segmentation-based models like YOLACT or Mask-RCNN. When converting to COCO format, both a training set and a validation set are created. For cases involving multiple animal pose estimation, the same user interface can be used to label the key points, which can then be converted into small regions of masks. The segmentation network treats pose estimation as a specialized use case. There are no limitations on what a user can label. In the most extreme case, every pixel in the video frame could be labeled. Our approach, referred to as Annolid, consists of four stages: preprocessing, labeling, training, motion tracking, and behavior analysis. To evaluate the efficacy of Annolid for segmenting instances, we meticulously labeled 200 frames randomly selected from a video featuring three interacting mice. The datasets provide annotations for combined instance masks and identity information, represented as class plus ID numbers (e.g., LeftMouse, RightMouse, and Mouse), as illustrated in Figure \ref{fig:three_mice}. Our trained models exhibit robust performance across diverse scenarios, effectively handling challenges such as multiple animals in rapid motion, occlusion at varying locations, and temporal variations from the camera. Notably, the model achieved an impressive average precision (AP) of 44.65 using 80 labeled frames in the training set. Furthermore, we extended our efforts to curate a Multi-Animal Tracking Dataset (MATD) 2023. This dataset comprises ten individual videos, each presenting challenging scenarios involving fast motion, occlusion, and other complex visual factors. Notably, the average length of each video in MATD 2023 exceeds 1000 frames, surpassing the maximum video duration observed in previous Video Object Segmentation (VOS) benchmarks. The datasets are available upon request.

\subsection{Training Instance Segmentation Models}
We applied a standard transfer learning strategy using the pre-trained weights from a network trained on the COCO datasets \cite{lin2014microsoft}.We fine-tuned the network on custom-labeled datasets. The YOLACT models were trained on a computer with an NVIDIA 1080Ti GPU card. The operating system was Windows 10 with CUDA 10.0, CUDNN 7.0, and PyTorch 1.4.0. With a single GPU on this machine, it takes a few hours for several thousand iterations. Inferencing on new videos is about 30 FPS in real-time on the trained model on the same GPU. We also provided Google Colab notebooks for training models. In addition to the strong performance, the approach is very simple and runs at 30 FPS on a single NVIDIA GTX 1080Ti GPU. It helps the scientists to adopt developing software by showing examples with neural networks. The input image is resized to 512 x 512, and the training step takes about 2 hours on one GTX 1080 Ti GPU. For training Mask-RCNN in Google Colab, we used Resnet50 as our default backbone with the model parameters pre-trained on the ImageNet \cite{deng2009imagenet} and COCO dataset \cite{lin2014microsoft} were used to initialize our model. We trained our model with the default optimizer implemented in Detectron2 for 3000 epochs with a starting learning rate of 0.00025. The batch size is set to 8. Other parameters were unchanged as the default configurations for Mask-RCNN FPN 3x as described in Detectron2 \cite{wu2019detectron2}.
 
 \section{Inferences}
During inference, we performed threshold-based filtering on the class scores predicted by the network. We did not need to follow the standard tracking methods to associate across frames. In our case, in each frame, the model was trained to predict the locations and identities of the objects.

\subsection{Evaluation}
In the evaluation of tracking accuracy for the test set within the two voles dataset, we employed the Multiple Object Tracking Accuracy (MOTA) metric \cite{bernardin2006multiple}. The dataset encompasses a total of 10,575 frames within the video. These results stem from predictions made by a model trained on a subset of 368 labeled frames.

The MOTA metric is defined as follows:
\begin{equation}
    MOTA = (1 - \frac{False Negatives + ID Switches + False Positives}{N_{GT}})
\end{equation}.

In our evaluation, we observed 52 instances of ID switches, 38 False Negative frames, and 11 False Positive frames. Applying the MOTA formula, the calculated MOTA score for this test video is 99.04\%.
We conducted an experiment using varying numbers of labeled frames in the training set, as depicted in Figure \ref{fig:tracking_examples_prediction}. To evaluate the performance of instance segmentation, we employed Average Precision (AP).

\begin{figure}
    \centering
    \includegraphics[scale=0.5]{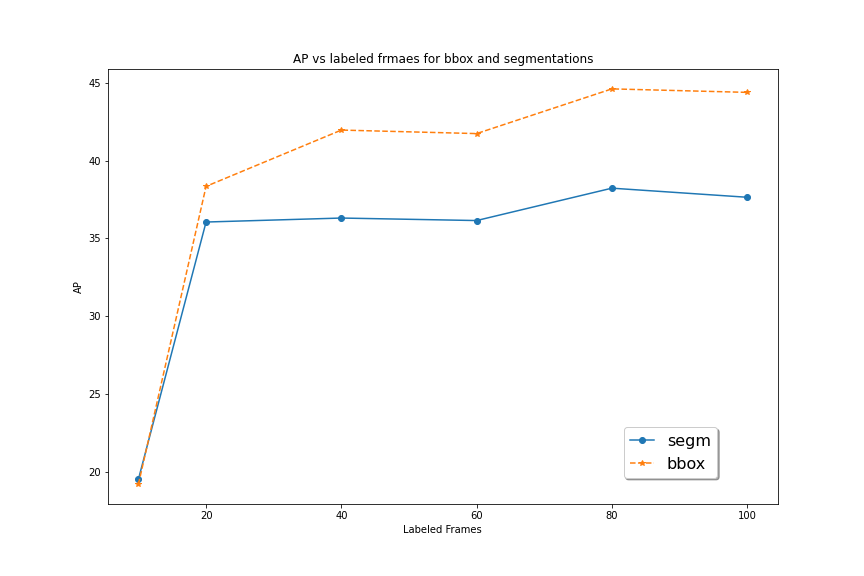}
    \caption{AP increases by training with more labeled frames}
    \label{fig:my_label}
\end{figure}

\subsection{Results}
In this section, we present the results on the Multi-Animal Tracking Dataset (MATD) 2023 in table \ref{tab:methods}. Since tracking based on instance segmentation can be seen as the classification of instance masks, we directly apply the AP measurements to evaluate the trained instance model.

\begin{table}[htbp]
  \centering
  \caption{Results on instance segmentation using the multi-animal tracking validation
dataset.}
    \begin{tabular}{|l|c|c|c|c|c|c|}
    \hline
          & \textbf{AP} $\uparrow$ & \textbf{AP50} $\uparrow$ & \textbf{AP75} $\uparrow$ & \textbf{APS} $\uparrow$ & \textbf{APM} $\uparrow$ & \textbf{APL} $\uparrow$ \\
    \hline
    \textbf{birds} & 64.735 & 86.246 & 75.953 & - & 39.360 & 65.971 \\
    \textbf{fish} & 65.806 & 96.153 & 82.054 & - & 58.408 & 73.087 \\
    \textbf{mice} & 79.906 & 97.006 & 93.182 & - & 79.906 & 0.0 \\
    \textbf{frogs} & 30.573 & 66.584 & 21.457 & 27.287 & 42.167 & - \\
    \textbf{voles} & 51.711 & 94.442 & 43.285 & 49.414 & 62.465 & - \\
    \textbf{polar bears} & 66.146 & 73.102 & 73.102 & - & 84.950 & 67.426 \\
    \textbf{ants} & 36.121 & 92.616 & 12.404 & 44.455 & 36.429 & - \\
    \textbf{bees} & 69.530 & 87.624 & 87.624 & - & - & 69.530 \\
     \textbf{white mice} & 69.328 & 71.926 & 71.926 & - & - & 69.328 \\
     \textbf{gerbils} & 66.349 & 90.064 & 71.211 & - & 68.735 & 68.298 \\
    \hline
    \end{tabular}%
  \label{tab:methods}%
\end{table}%

\begin{figure}[h]
    \centering
    \includegraphics[width=\linewidth]{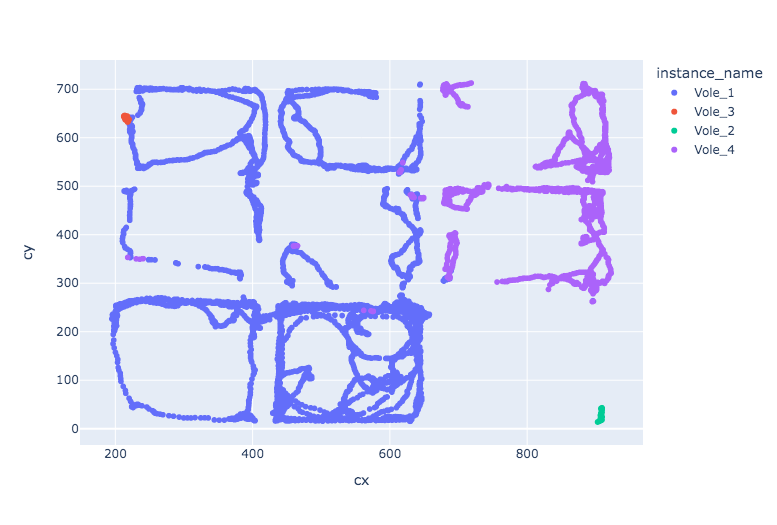}
    \caption{Overview of tracks for four voles in a 6-minute stitched video.}
    \label{fig:vole_tracks}
\end{figure}

\begin{figure}
    \centering
    \includegraphics[width=\textwidth,height=\textheight,keepaspectratio]{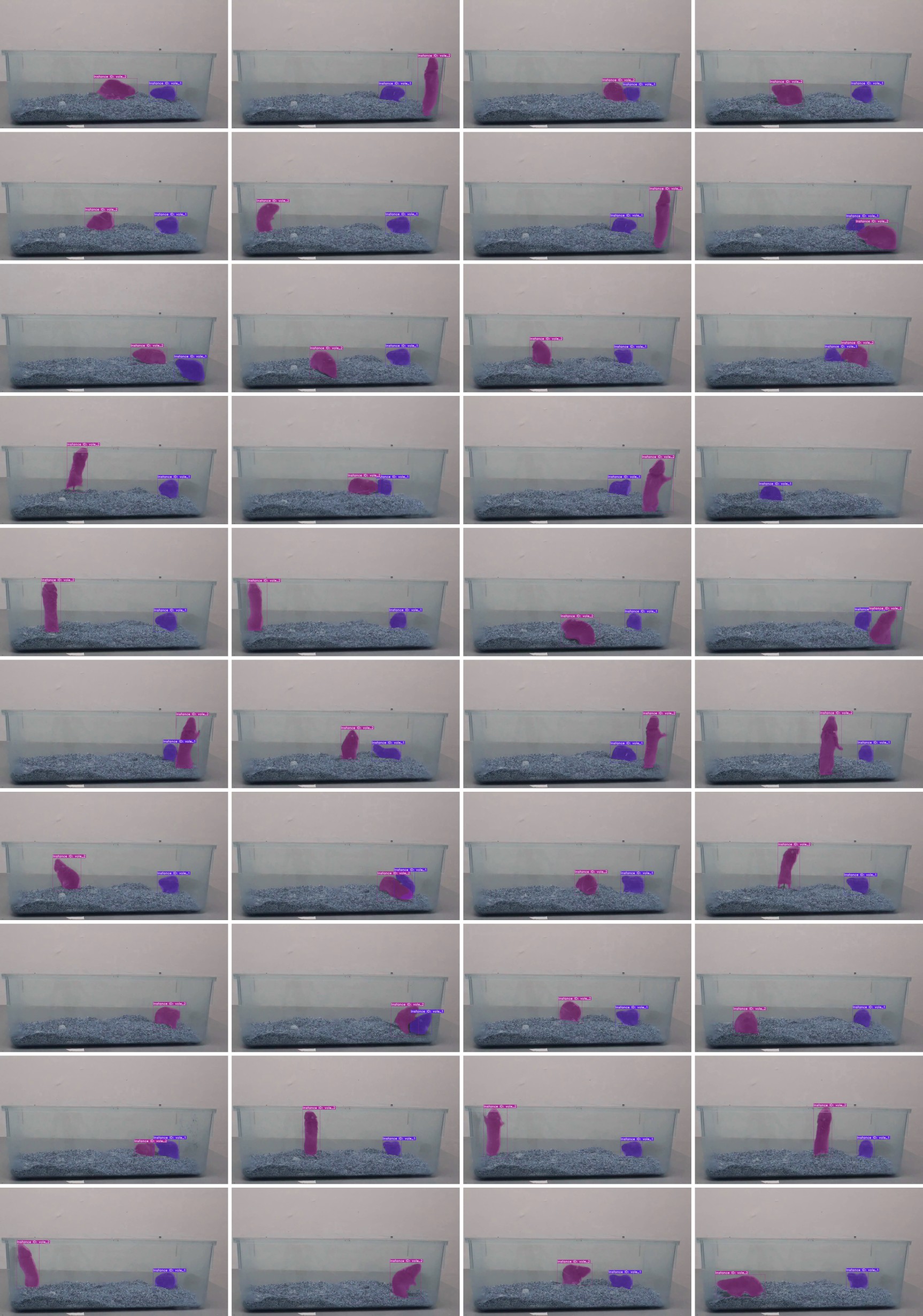}
    \caption{Randomly sampled tracking results of Annolid for two interacting voles, with color-coded masks and bounding boxes. For better visibility, zoom in to see the colors in detail.}
    \label{fig:tracking_vole_examples}
\end{figure}

In this study, we present qualitative results obtained from the two voles dataset, visualized in Figure \ref{fig:tracking_vole_examples}. Our proposed approach exhibits remarkable performance, notably reducing the occurrence of ID switches. Moreover, our method achieves inference speeds approaching real-time, especially noteworthy for YOLACT models. These results are compelling, given the simplicity of our approach, which refrains from introducing complexity for frame-to-frame association. Furthermore, we conducted a comprehensive evaluation of Annolid's performance across varying amounts of labeled training data. Impressively, utilizing approximately 100 labeled frames for training yielded high Multiple Object Tracking Accuracy (MOTA), approaching human-level performance. In a test video featuring two voles and comprising a total of 10,575 frames, we computed 52 ID switches (0.49\% of frames), 38 false negatives (0.36\%), and 11 false positives (0.1\%). The resulting MOTA score reached an impressive 99.05\%. Figure \ref{fig:tracking_vole_examples} showcases 40 randomly selected tracking frames, with each row presenting sampled frames in chronological order. The images are annotated with segmentation masks and identities, with distinct colors representing individual animals. For optimal visual appreciation, we recommend viewing the figure in color.

\subsection{Qualitative Results}
\begin{figure}
    \centering
    \includegraphics[width=\textwidth,height=\textheight,keepaspectratio]{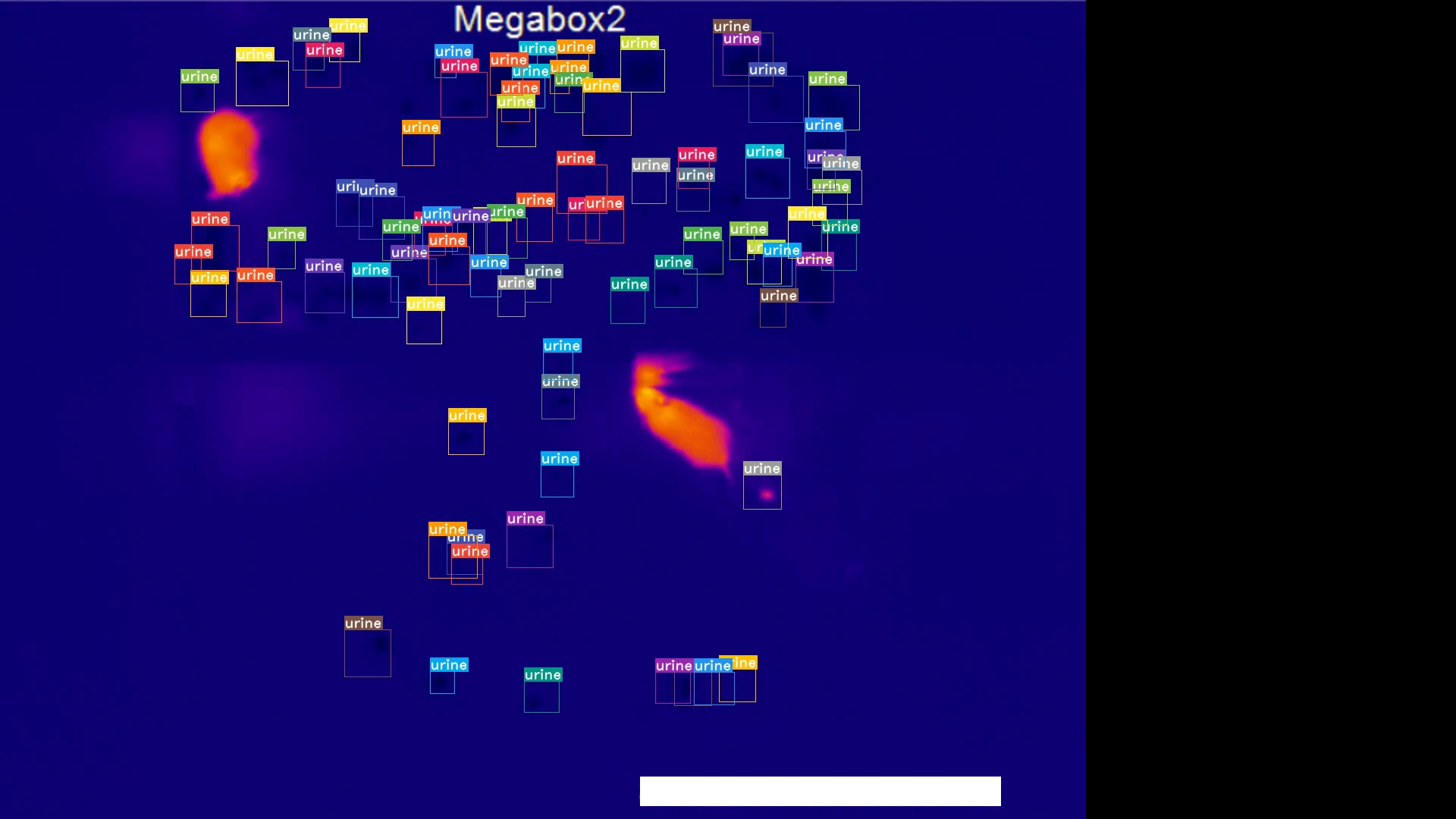}
    \caption{An example of detected urine spots, displayed with bounding boxes.}
    \label{fig:tracking_examples_prediction}
\end{figure}

In this section, we present a selection of qualitative results from the tracking process. For additional use cases and more examples, please refer to the appendix. Figure \ref{fig:pose_estimation} showcases exemplary results of our pose estimation, highlighting precise tracking of specific body parts such as the nose, left ear, right ear, body centroid, and tail base. Annolid demonstrates successful application across diverse experimental videos, showcasing its capability to track segmented regions of interest according to users' specific needs and expectations.Based on the results shown in Figure \ref{fig:tracking_examples_prediction} and Figure \ref{fig:tracking_examples}, our method successfully demonstrates instance segmentation and recognition even in complex scenarios such as two animals crossing over each other, while effectively recovering from occlusions.


\section{Discussion}
In our implementation, we embrace a multiple-object tracking classification approach that extracts information solely from the current frames, distinct from batch methods that rely on future frames. Online methodologies, exemplified by SORT \cite{Wojke2017simple} and DeepSort \cite{Wojke2018deep}, address the data association challenge by deploying techniques such as the Kalman Filter \cite{welch1995introduction} for object motion prediction and the Hungarian algorithm \cite{kuhn1955hungarian} for data association. Notably, our work departs from utilizing motion information derived from adjacent frames to enhance tracking accuracy. In contrast, Hou et al. \cite{hou2017tube} introduced a Tube Convolutional Neural Network (T-CNN) in a study, leveraging 3D convolution features in equal-length video clips for action recognition and localization. While we don't incorporate motion information from adjacent frames in our current work, our future exploration involves integrating networks like T-CNN. This investigation aims to uncover potential advantages, particularly in the realm of self-supervised learning. In our envisioned future work, we plan to delve into the integration of networks such as T-CNN and explore the potential benefits of self-supervised learning. This approach holds promise for reducing annotation efforts, rendering Annolid more attractive for diverse applications, especially in scenarios where behavior is influenced by disease or genetic engineering. The use of self-supervised learning enables investigators to circumvent bias introduced through manual labeling, allowing the model to learn behavior directly through instance segmentation.

\section{Conclusion}
In this paper, we present our instance segmentation approach, which effectively improves tracking accuracy without requiring extensive engineering efforts. The results demonstrate that our approach offers a significant advantage in behavioral experiments by not being heavily reliant on labeled data. We introduce Annolid, a framework that achieves comparable performance to pose estimation and multiple object tracking (MOT) methods on our benchmark datasets, in terms of both tracking accuracy and inference speed. Furthermore, Annolid inherently exhibits data-efficiency by leveraging network overfitting capabilities, making it highly valuable for applications involving multiple animal tracking and behavior analysis.

\section{Acknowledgements}
This work was supported by NSF grant 1743214.
\vspace{2in}
\bibliographystyle{\style}
\bibliography{citations}

\newpage
\appendix
\section{Appendix}

\subsection{Detect deposited urine}

\begin{figure}[!ht]
    \centering
    \includegraphics[width=\textwidth,height=\textheight,keepaspectratio]{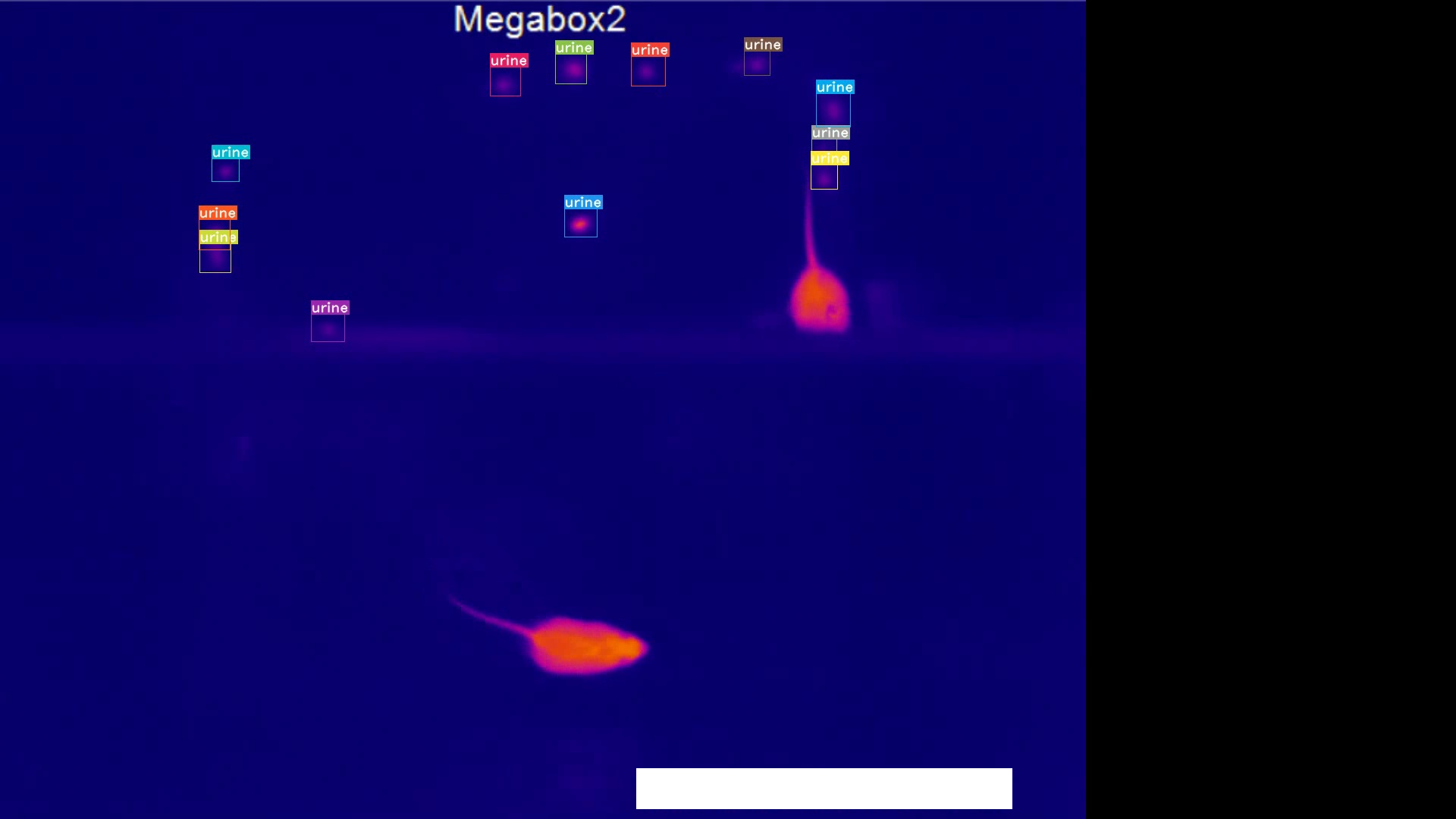}
    \caption{An example of detected newly deposited urine display, highlighted with bounding boxes.}
    \label{fig:tracking_examples}
\end{figure}

\subsection{Tracking results of Annolid for three interacting voles and their body parts}

\begin{figure}[!ht]
    \centering
    \includegraphics[width=\textwidth,height=\textheight]{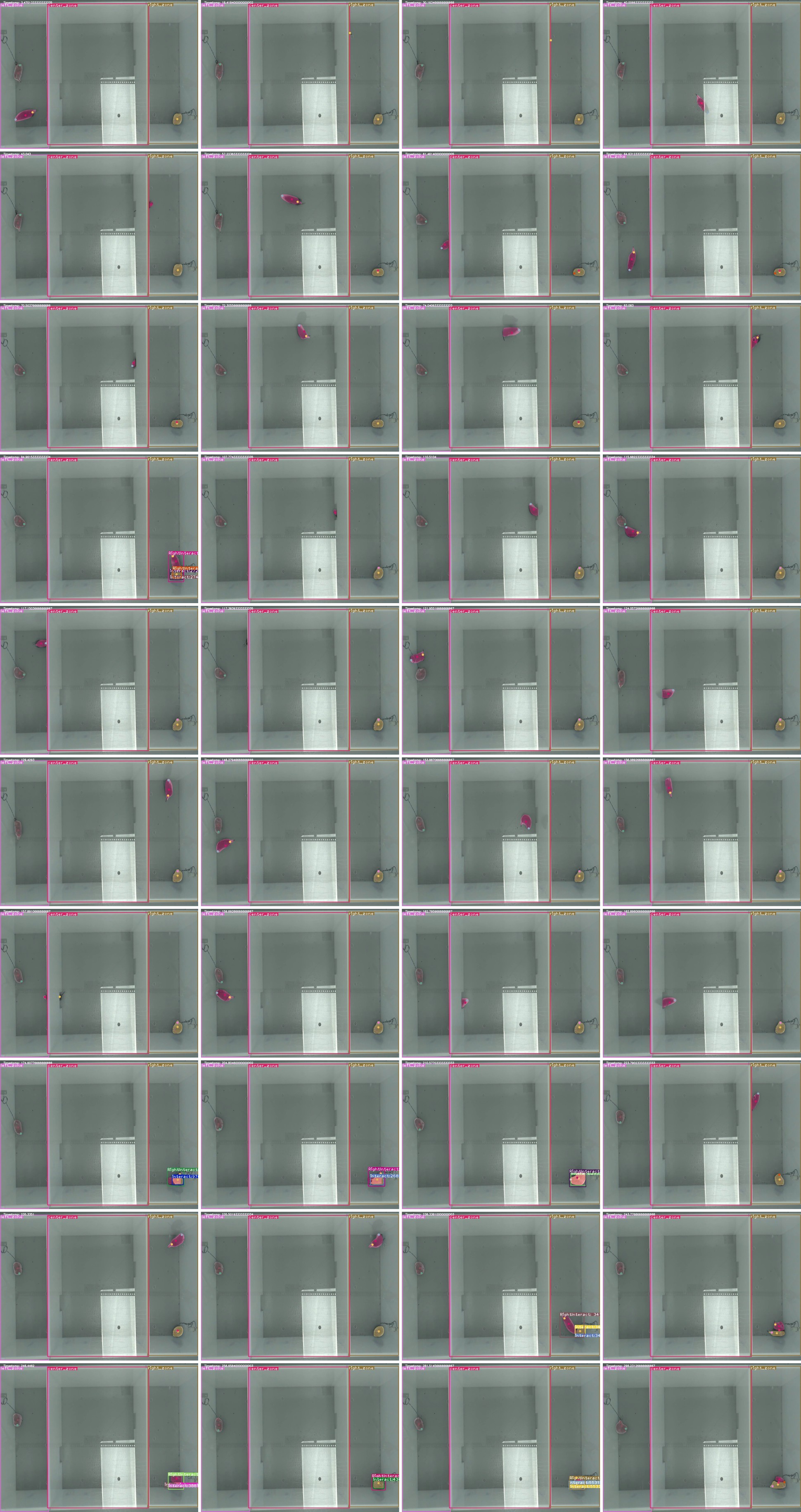}
    \caption{Randomly sampled tracking results of Annolid for three interacting voles and their body parts, with color-coded masks. For optimal viewing, zoom in to see the colors and details.}
    \label{fig:tracking_three_voles_examples}
\end{figure}
\end{document}